\documentclass{article} 

\usepackage{hyperref}
\usepackage{url}
\usepackage{enumitem}
\usepackage{booktabs}   
\usepackage{multirow}   
\usepackage{amsmath} 
\usepackage{xspace}
\usepackage[capitalize,noabbrev]{cleveref}
\usepackage{graphicx}   
\usepackage{subcaption} 
\usepackage{caption}
\usepackage{listings}
\usepackage[most]{tcolorbox}
\usepackage{xcolor} 
\usepackage[ruled,vlined,linesnumbered]{algorithm2e}
\usepackage{wrapfig}
\usepackage{calc}
\usepackage{makecell}

\definecolor{lightgray}{RGB}{245,245,245}

\crefformat{section}{\S#2#1#3}
\crefformat{subsection}{\S#2#1#3}
\crefformat{subsubsection}{\S#2#1#3}
\crefrangeformat{section}{\S#3#1#4 to~\S#5#2#6}
\crefmultiformat{section}{\S#2#1#3}{ and~\S#2#1#3}{, #2#1#3}{ and~#2#1#3}
\Crefformat{figure}{#2Figure~#1#3}
\Crefmultiformat{figure}{Figs.~#2#1#3}{ and~#2#1#3}{, #2#1#3}{ and~#2#1#3}
\Crefformat{table}{#2Table~#1#3}
\Crefmultiformat{table}{Tabs.~#2#1#3}{ and~#2#1#3}{, #2#1#3}{ and~#2#1#3}
\Crefformat{appendix}{#2Appx.~\S#1#3}
\crefformat{algorithm}{Alg.~#2#1#3}
\Crefformat{algorithm}{Alg.~#2#1#3}
\crefname{algocf}{Alg.}{Algs.}
\Crefname{algocf}{Alg.}{Algs.}
\Crefformat{equation}{#2Eq.~#1#3}

\newcommand{\sys}{\mbox{\textsc{FutureWeaver}}\xspace}

\usepackage[preprint]{colm2026_conference}

\usepackage{microtype}
\usepackage{hyperref}
\usepackage{url}
\usepackage{booktabs}

\usepackage{lineno}

\newcommand{\stitle}[1]{\vspace{0.5em}\noindent{\bf #1}}

\definecolor{darkblue}{rgb}{0, 0, 0.5}
\hypersetup{colorlinks=true, citecolor=darkblue, linkcolor=darkblue, urlcolor=darkblue}

\title{\sys: Planning Test-Time Compute for Multi-Agent Systems with Modular Collaboration}

\author{
  Dongwon Jung\textsuperscript{1}  \quad 
  Peng Shi\textsuperscript{2}   \quad 
  Muhao Chen\textsuperscript{1}    \quad 
  Yi Zhang\textsuperscript{3} \\
  \textsuperscript{1}University of California, Davis \quad
  \textsuperscript{2}University of Waterloo \quad
  \textsuperscript{3}Greenshoe, Inc. 
}

\begin{document}

\ifcolmsubmission
\linenumbers
\fi

\maketitle

\begin{abstract}
Scaling test-time computation has been shown to significantly improve large language model (LLM) performance without additional training.
However, extending these techniques to multi-agent systems remains challenging: existing approaches lack principled mechanisms for allocating compute to enable effective collaboration, scaling coordination itself, or optimizing compute usage under explicit budget constraints.
To address this gap, we propose \sys, a framework for \emph{planning and optimizing} test-time compute allocation in multi-agent systems under fixed budgets.
It introduces \emph{collaboration modules}, formalized as modular, callable functions that encapsulate reusable multi-agent workflows and are automatically induced via self-play reflection from recurring interaction patterns.
Building on these modules, it employs \emph{a dual-level planning architecture} that jointly performs short-horizon action selection and long-horizon abstract lookahead to optimize inference trajectories under budget constraints. Experiments on complex agent benchmarks demonstrate that \sys consistently outperforms baselines across diverse budget settings, validating its effectiveness for multi-agent collaboration in inference-time optimization.
\end{abstract}

\section{Introduction}

Increasing inference-time computation~\citep{snell2024scaling,s1-2025,balachandran2025inference,zhang2025survey} has emerged as a powerful strategy for improving the performance of large language models (LLMs) without additional training. Recent reasoning-oriented models, such as OpenAI’s o-series and Anthropic’s Claude models, have demonstrated strong performance gains on reasoning through techniques such as self-correction~\citep{self-refine2023,chen2025sets}, iterative verification~\citep{lee2025evolving}, and best-of-N sampling~\citep{brown2024large}. However, extending test-time scaling to multi-agent systems~\citep{openai2025deepresearch,google2025gemini,anthropic2024claudecode,openai2024codex} introduces unique challenges.

First, standard test-time scaling techniques for LLMs do not extend cleanly to multi-agent systems. In single-agent settings, additional compute is applied within a single model invocation, e.g., longer reasoning traces, repeated sampling, or self-verification.
In multi-agent settings, however, scaling is fundamentally a trajectory-level \emph{collaboration} problem: compute must be \emph{allocated} across heterogeneous interactions, deciding which agents to invoke, when to verify, and how to integrate intermediate artifacts.
The widely adopted orchestrator-worker paradigm \citep{hadfield2025multiagent,tran2025multi} largely executes agents sequentially and aggregates outputs, offering limited mechanisms for scaling coordination itself.
Consequently, systems often fall back to rigid calling routines and leave complementarities between agents underexploited. 


Second, even with growing interest in budget-aware test-time scaling, there remains no principled mechanism for constructing the most effective inference trajectory under a hard compute budget. Recent methods regulate test-time cost through budget forcing, e.g., s1~\citep{s1-2025}, or adaptive decomposition across sub-questions, e.g., Plan-and-Budget~\citep{lin2025plan}. However, these approaches largely optimize within a fixed inference structure, such as a single agent reasoning or a predetermined decomposition, rather than treating inference as a trajectory \emph{planning} problem over a combinatorial space of agent invocations and coordination steps. Rather than merely ensuring that an inference trajectory stays within budget, an effective system should explicitly select the most effective sequence of agent invocations and coordination steps that the budget can afford. This distinction is critical: the benefit of spending compute at each step depends on the downstream trajectory it enables. For instance, systems might over-invest in early, low-leverage steps or under-invest when later steps demand heavier reasoning or richer coordination. The challenge is further amplified in multi-agent systems, where the optimal use of a budget often corresponds to a particular ordering of agent and module invocations with heterogeneous costs and capabilities, rather than simply allocating more tokens or repeats.


In this paper, we address the core question: \textit{How can multi-agent systems maximize performance under fixed compute budgets while effectively leveraging complementary agents through collaboration?} To this end, we propose \sys, a general framework for budget-constrained optimization of test-time compute in multi-agent systems. 
At the core of \sys is the abstraction of \emph{collaboration modules}, which are modular, callable functions that encapsulate high-level coordination strategies among agents. 
Each module specifies a structured multi-agent workflow with standardized inputs and functionality, rendering agent interactions composable and reusable, akin to tools in function-calling frameworks. This abstraction enables principled allocation and scaling of compute, as the orchestrator can dynamically decide when and how to invoke increasingly compute-intensive forms of collaboration based on the evolving task state and remaining budget. From this perspective, multi-agent coordination is cast as a function-calling problem, with an orchestrator agent adaptively choosing among collaboration modules. The modules themselves are automatically induced via self-play reflection over past trajectories, capturing reusable coordination patterns without manual design.

Having established reusable collaboration modules, the remaining challenge is to optimize computation allocation across them under fixed budget constraints.
To address this, \sys employs a \emph{dual-level planning architecture} that integrates short-horizon and long-horizon planning. 
The short-horizon planner proposes and ranks candidate next actions, either invoking a collaboration module or an individual agent, conditioned on the current task state. 
In parallel, the long-horizon planner, inspired by speculative decoding \citep{speculativedecoding2023}, performs abstract lookahead by reasoning over potential sequences of collaboration modules without executing them. 
This yields low-cost estimates of future compute consumption and budget feasibility, which are fed back to guide the short-horizon planner toward actions that lead to the most effective compute-feasible inference trajectories.
Rather than serving solely as a budget constraint check, long-horizon planning actively shapes local decisions by anticipating future coordination demands.  Together, the two levels operate in an A*-like manner, jointly optimizing over the combinatorial space of multi-agent collaboration trajectories under strict budget constraints.

To summarize, this paper makes the following contributions:
\begin{itemize}[leftmargin=*]
\item We formulate the problem of optimizing test-time compute allocation for multi-agent systems under fixed budget constraints with a set of agents capable of collaboration.
\item We propose \sys, a framework combining collaboration modules, dual-level planning architecture, and self-play reflection for budget-aware multi-agent collaboration.
\item We demonstrate consistent improvements over baselines across challenging multi-agent benchmarks in both task success and budget utilization.
\end{itemize}
The remaining sections formalize the problem setup (\Cref{sec:2}), present our method (\Cref{sec:3}), and provide experimental evaluation and analysis (\Cref{sec:4}). Related work is discussed in \Cref{sec:related_work}.

\section{Orchestrator--Worker Framework Under Budget Constraints}
\label{sec:2}
We now formalize the orchestrator–worker framework  \citep{hadfield2025multiagent,tran2025multi} for multi-agent systems operating under a fixed test-time compute budget.

Let $\mathcal{A}=\{a_1,\dots,a_N\}$ denote the set of available \textit{worker agents}, each initialized with a large language model (LLM) with distinct capabilities and an associated cost of invocation. A task is solved through a sequence of intermediate states $\{s_t\}_{t=0}^{H}$, where $s_0$ is the initial task description and $s_H$ is the final output returned by the system. At each step $t$, the orchestrator selects an \textit{action} $\alpha_{\scriptscriptstyle t}=(a_{\scriptscriptstyle t},\upsilon_{\scriptscriptstyle t})$,
where $a_{\scriptscriptstyle t}\in\mathcal{A}$ and $\upsilon_{\scriptscriptstyle t}$ specifies the subtask or input to the chosen worker. Executing $\alpha_{\scriptscriptstyle t}$ produces an output $o_{\scriptscriptstyle t}$, and the new state is defined as
$s_{\scriptscriptstyle t} = (\upsilon_{\scriptscriptstyle t},\, o_{\scriptscriptstyle t})$.
Each action incurs a cost $\mathrm{cost}(\alpha_{\scriptscriptstyle t})$, and the cumulative cost is constrained by the budget $B$, i.e.
$
\sum_{t=1}^{H}\mathrm{cost}(\alpha_{\scriptscriptstyle t}) \le B.
$
The system's behavior at step $t$ is governed by a policy $\pi$ that selects the next action based on the execution history
$\mathcal{H}_t=\{\,s_r\,\}_{r=0}^{t-1}$.

We instantiate the policy $\pi$ with an LLM \citep{yao2020keep,huang2022language,yao2023react}, which serves as the backbone for the orchestrator agent. Following the ReAct framework \citep{yao2023react}, the orchestrator generates subtasks, assigns them to worker agents, evaluates their outputs, and determines subsequent actions until a final solution is produced or the budget is exhausted.

\begin{figure*}[t]
    \centering
\small\includegraphics[width=0.9\linewidth]{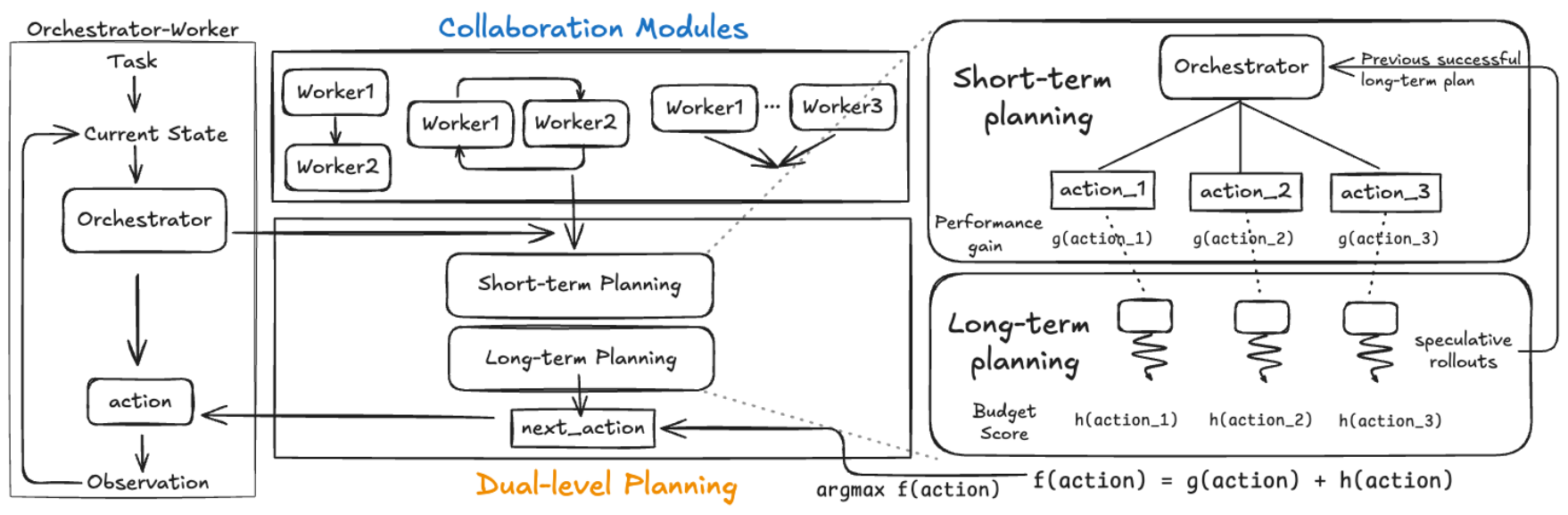}
    \caption{Overview of \sys. The framework extends the standard orchestrator–worker paradigm by introducing \textit{collaboration modules} and \textit{dual-level planning architecture}.}

    \label{fig:method}
\vspace{-1em}
\end{figure*}
\section{\sys}
\label{sec:3}
Multi-agent systems built on the orchestrator–worker paradigm \citep{hadfield2025multiagent,tran2025multi} suffer from limited coordination and inefficient use of additional compute. Synergies between agents are underexploited, and budget-aware optimization has not been well studied. 
To address these issues, we propose \sys, which extends the orchestrator–worker framework with three key components: Collaboration Modules (\Cref{sec:collaboration_modules}), Dual-Level Planning Architecture (\Cref{sec:dual_level_planning}), and Self-Play Reflection (\Cref{sec:self_play_reflection}). \Cref{fig:method} illustrates the overview of our method.




\subsection{Collaboration Modules}
\label{sec:collaboration_modules}

The traditional orchestrator–worker framework is constrained by sequential, one-agent-at-a-time execution. This design becomes inefficient on complex, open-ended tasks, as it is difficult to capture synergies between agents with complementary capabilities. In particular, the sequential setup limits coordination, since intermediate results are not jointly integrated across agents, leaving subtasks only partially addressed. Moreover, this rigid design restricts how additional compute can be used: allocating more resources to a single agent, through repeated sampling or iterative verification, merely amplifies that agent’s behavior without fostering cross-agent collaboration. Unlike test-time scaling in single-agent LLMs, such additional compute does not provide a principled path to improving performance in multi-agent systems.

To address these limitations, we introduce \textit{collaboration modules}, modular and callable functions that encapsulate high-level coordination strategies among multiple agents. Each module specifies a standardized and reusable workflow, such as combining outputs from multiple agents or chaining agents in a pipeline, and can be invoked with a single function call. This abstraction reframes multi-agent collaboration as a function-calling problem, where the orchestrator decides which collaboration module to invoke at each step. By structuring interactions in this way, collaboration modules enable richer coordination and provide a scalable and principled mechanism for utilizing test-time compute more effectively, as additional budget is allocated to cross-agent collaboration rather than simply amplifying the behavior of individual agents.

Formally, we define a collaboration module as $m = (\mathcal{S}, \kappa)$
where $\mathcal{S}\subseteq \mathcal{A}$ denotes a subset of worker agents and $\kappa$ is a coordination strategy that specifies how these agents interact. The action space available to the orchestrator is therefore expanded to include not only the individual base agents but also collaboration modules defined as, $\mathcal{A}' = \mathcal{A} \cup \mathcal{M}$,
where $\mathcal{A}$ is the set of base worker agents and $\mathcal{M}$ is the set of collaboration modules. From this point onward, we assume that $a_{\scriptscriptstyle t}$ in action $\alpha_{\scriptscriptstyle t}$ is drawn from $\mathcal{A}'$, enabling the orchestrator to invoke either a single agent or a collaboration module at each step.
 
\subsection{Dual-Level Planning Architecture}
\label{sec:dual_level_planning}
While collaboration modules provide a structured abstraction for multi-agent coordination, they also introduce the challenge of choosing, among heterogeneous agent/module invocations at each step with different costs and downstream implications. Under a hard budget, distributing test-time resources across modules is inherently uncertain, since the system cannot know a priori which coordination choices will be needed later or how costly future interactions will become.
We therefore require an inference-time planner that jointly prioritizes promising immediate actions and anticipates whether they can be extended into effective, budget-feasible trajectories.

Inspired by best-first search and A*~\citep{hart1968formal,meng2024llm}, we treat inference as an iterative expansion of a search tree over actions $\alpha_{\scriptscriptstyle t} = (a_{\scriptscriptstyle t}, \upsilon_{\scriptscriptstyle t})$, where $a_{\scriptscriptstyle t}$ denotes an agent or collaboration module and $\upsilon_{\scriptscriptstyle t}$ is its instantiated input.\footnote{We use A* only as an organizing analogy: $g(\cdot)$ captures local progress signals, while $h(\cdot)$ provides a low-cost lookahead signal over remaining budget feasibility.} Each candidate action is scored by two complementary terms: a short-horizon gain $g(\alpha_{\scriptscriptstyle t})$ that ranks immediate promise given the current history $\mathcal{H}_t$, and a long-horizon feasibility score $h(\alpha_{\scriptscriptstyle t})$ that estimates whether $\alpha_{\scriptscriptstyle t}$ admits plausible budget-compliant continuations. Crucially, the feasible speculative set produced by long-horizon planning is carried forward to condition the next step's short-horizon proposal distribution, coupling lookahead with candidate generation rather than using feasibility as a passive filter.


By combining these two signals, \sys selects the action with the highest overall utility
$f(\alpha_{\scriptscriptstyle t}) = g(\alpha_{\scriptscriptstyle t}) + h(\alpha_{\scriptscriptstyle t}),$
ensuring that each immediate step is not only locally promising but also aligned with a globally feasible, budget-respecting trajectory. 
This design is complementary to query-level workflow synthesis methods that generate a multi-agent workflow per query~\citep{gao2025flowreasoner}, and to budget-aware system structuring approaches that select topologies prior to execution~\citep{Yang2025BAMASSB}. In contrast, \sys performs step-wise planning during inference under a fixed budget, using low-overhead module-level lookahead to guide which collaboration to execute next. It is also distinct from inference-time search for agents that relies on environment-level rollouts~\citep{Koh2024TreeSF}, as our lookahead operates symbolically over collaboration modules with costs estimated from past experience, avoiding expensive execution during speculation.

\subsubsection{Short-Term Planning}
Short-term planning evaluates the immediate utility of a candidate action $\alpha_{\scriptscriptstyle t} = (a_{\scriptscriptstyle t}, \upsilon_{\scriptscriptstyle t})$ proposed by a policy $\pi$. 
At each step $t$, the policy samples $K$ candidates
\[
\mathcal{C}_t = \{\alpha_{\scriptscriptstyle t}^{\scriptscriptstyle(1)} , \dots, \alpha_{\scriptscriptstyle t}^{\scriptscriptstyle(K)} \} \sim \pi(\cdot \mid \mathcal{H}_t, \mathcal{T}_{\text{feasible},t-1}),
\]
where $\mathcal{T}_{\text{feasible},t-1}$ is the set of feasible speculative trajectories returned by the previous long-horizon planning step. Conditioning on $\mathcal{T}_{\text{feasible},t-1}$ biases proposals toward actions compatible with budget-feasible high-level plans. We define the short-horizon gain $g(\alpha_{\scriptscriptstyle t}^{\scriptscriptstyle(i)})$ by combining two signals.

(1) \textbf{Self-Consistency ($SC$)} measures local consensus among sampled actions as a proxy for confidence. Let $\phi((a,v))=a$ extract the agent or module from an action. Then
\begin{equation*}
    SC(\alpha_{\scriptscriptstyle t}^{\scriptscriptstyle(i)}) = \frac{1}{K}\sum_{k=1}^{K} \mathbf{1}\!\left[\phi(\alpha_{\scriptscriptstyle t}^{\scriptscriptstyle(k)})=\phi(\alpha_{\scriptscriptstyle t}^{\scriptscriptstyle(i)})\right]
\end{equation*}
where $\mathbf{1}[\cdot]$ is the indicator function. A higher $SC$ indicates stronger confidence that a specific component will contribute effectively to task progress.

(2) \textbf{Performance Prior ($PP$)} captures the empirical reliability of a collaboration module or agent by serving as a trajectory-level performance prior. It is derived at inference time by retrieving and aggregating the average success rates of similar trajectories from past experience collected through self-play reflection (Section~\ref{sec:self_play_reflection}). By grounding the estimate in outcomes of complete execution trajectories, $PP$ reflects not only immediate effectiveness but also the long-term contribution of a component to successful budget-feasible plans in comparable task contexts.

Finally, the performance gain $g(\alpha_{\scriptscriptstyle t}^{\scriptscriptstyle(i)})$ integrates these signals using a geometric mean, ensuring an action is favored only if it is both contextually relevant and empirically effective:
\begin{equation*}
    g(\alpha_{t}^{\scriptscriptstyle(i)}) = \sqrt{ SC(\alpha_{t}^{\scriptscriptstyle(i)}) \cdot PP(\alpha_{t}^{\scriptscriptstyle(i)}) }.
\end{equation*}
It penalizes actions lacking either local consensus or historical reliability, providing a stable signal for prioritizing near-term steps before long-term feasibility is assessed.


\subsubsection{Long-Term Planning}

While short-term planning evaluates the immediate promise of an action, long-term planning speculates on whether that action can lead to a high-performing trajectory that remains within the remaining budget. The goal is to compute a feasibility score $h(\alpha_{\scriptscriptstyle t}^{\scriptscriptstyle(i)})$ that reflects how likely it is that choosing $\alpha_{\scriptscriptstyle t}^{\scriptscriptstyle(i)}$ will keep future steps budget-compliant while still allowing the system to pursue high-utility continuations -- all without executing $\alpha_{\scriptscriptstyle t}^{\scriptscriptstyle(i)}$ or any subsequent actions.

Concretely, the orchestrator agent expands each candidate action into \emph{speculative trajectories}, which are abstract rollouts representing possible continuations of agents or collaboration modules. 
These trajectories are generated symbolically at the module level, so no agents  are invoked, keeping the procedure lightweight. Each trajectory is associated with an estimated cumulative cost derived from Self-Play Reflection, and any that exceed the remaining budget are filtered out. 
Formally, let $\mathcal{T}_t(\alpha_{\scriptscriptstyle t}^{\scriptscriptstyle(i)})$ denote the set of speculative trajectories beginning with $\alpha_{\scriptscriptstyle t}^{\scriptscriptstyle(i)}$, and let $\mathrm{cost}(\tau)$ represent the estimated cumulative cost of a trajectory $\tau \in \mathcal{T}_t(\alpha_{\scriptscriptstyle t}^{\scriptscriptstyle(i)})$. Given the remaining budget $B_t$, we define the set of feasible trajectories as
\[
\mathcal{T}_{\text{feasible},t}(\alpha_{\scriptscriptstyle t}^{\scriptscriptstyle(i)}) = \{\tau \in \mathcal{T}_t(\alpha_{\scriptscriptstyle t}^{\scriptscriptstyle(i)}) \mid \mathrm{cost}(\tau) \leq B_t\},
\]
so that trajectories whose costs exceed $B_t$ are filtered out. Among the feasible trajectories, we then normalize across the $K$ candidate actions sampled at step $t$. This defines the budget feasibility score:
\[
h(\alpha_{\scriptscriptstyle t}^{\scriptscriptstyle(i)}) = \frac{|\mathcal{T}_{\text{feasible},t}(\alpha_{\scriptscriptstyle t}^{\scriptscriptstyle(i)})|}{\sum_{r=1}^{K} |\mathcal{T}_{\text{feasible},t}(\alpha_{\scriptscriptstyle t}^{(r)})|}.
\]
Intuitively, $h(\alpha_{\scriptscriptstyle t}^{\scriptscriptstyle(i)} )$ reflects the likelihood that an action can be extended into a successful budget-compliant plan. Actions with higher $h(\alpha_{\scriptscriptstyle t}^{\scriptscriptstyle(i)} )$ are favored, since they not only appear promising in the short term but also are more likely to sustain progress without violating budget constraints. This speculative lookahead ensures immediate choices remain consistent with long-term feasibility, complementing the performance gain $g(\alpha_{\scriptscriptstyle t}^{\scriptscriptstyle(i)} )$ in the overall utility $f(\alpha_{\scriptscriptstyle t}^{\scriptscriptstyle(i)} ) = g(\alpha_{\scriptscriptstyle t}^{\scriptscriptstyle(i)} ) + h(\alpha_{\scriptscriptstyle t}^{\scriptscriptstyle(i)} )$. Moreover, the resulting feasible speculative set $\mathcal{T}_{\text{feasible},t}$ is carried forward to guide candidate generation in the next step’s short-term planning.

Finally, we select the candidate action with the highest utility score as the next action. The dual-level planning procedure is formally presented in \Cref{alg:dual_planning}.

\subsection{Self-Play Reflection}
\label{sec:self_play_reflection}

Dual-level planning requires estimates of the performance and execution cost for agents and collaboration modules. Additionally, for collaboration modules, we require an effective way to structure multi-agent collaborations. To address both needs, \sys employs \textit{self-play reflection}, an offline, iterative process that builds experience by generating execution trajectories on a subset of validation tasks and uses this experience both to (1) collect performance and cost estimates of each agent or module and to (2) automatically construct collaboration modules from recurring interaction patterns.  

Formally, given a set of actions $\mathcal{A}'$, the system executes multiple trajectories on validation tasks to construct the experience buffer $\mathcal{D}_{\text{exp}}$, logging the agents invoked, subtasks addressed, outputs produced, and costs incurred. This dataset provides the necessary statistics for both planning levels. 

\textbf{Cost Estimation.} 
While the cost of invoking an agent or module theoretically depends on both the component $a$ and the specific subtask $\upsilon$, enumerating all possible subtasks is infeasible. We therefore relax this assumption and approximate the cost by averaging across observed subtasks in self-play trajectories, treating the resulting estimate as transferable across tasks\footnote{We provide an empirical analysis of the gap between cost estimation and actual costs in \Cref{sec:reliabilty_mean_cost}.}. Thus, the average execution cost of each agent or module is estimated as:
\begin{equation*}
    \widehat{\text{cost}}(a) = \mathbb{E}_{\tau \in \mathcal{D}_{\text{ref}}} [\text{cost}(a, v)], \quad a \in \mathcal{A}'
\end{equation*}
\textbf{Performance Estimation.} The performance estimation from $\mathcal{D}_{\text{exp}}$ serves as the performance prior in short-term planning, acting as the empirical prior for performance. To make this estimation context-aware, let $\psi(\alpha_{\scriptscriptstyle t}^{\scriptscriptstyle(i)} ) = \upsilon_{\scriptscriptstyle t}^{\scriptscriptstyle(i)} $ be a function that extracts the subquery from the candidate action $\alpha_{\scriptscriptstyle t}^{\scriptscriptstyle(i)} $. We then let $S_{\scriptscriptstyle\text{top10}}(\upsilon_{\scriptscriptstyle t}^{\scriptscriptstyle(i)} )$ denote the set of the top-10 subqueries in $\mathcal{D}_{\text{exp}}$ that are most semantically similar to $\upsilon_{\scriptscriptstyle t}^{\scriptscriptstyle(i)} $. We define $\mathcal{T}(S_{\scriptscriptstyle\text{top10}})$ as the set of unique trajectories in the experience buffer that contain at least one of these retrieved subqueries. The $PP$ is then calculated as the mean success rate across these trajectories:
\begin{equation*}
PP(\alpha_{\scriptscriptstyle t}^{\scriptscriptstyle(i)} ) = \frac{\sum_{\tau \in \mathcal{T}(S_{\scriptscriptstyle\text{top10}}(\psi(\alpha_{\scriptscriptstyle t}^{\scriptscriptstyle(i)} ))} \mathbf{1}[Success(\tau)]}{|\mathcal{T}(S_{\scriptscriptstyle\text{top10}}(\psi(\alpha_{\scriptscriptstyle t}^{\scriptscriptstyle(i)} ))|}
\end{equation*} 
where $Success(\tau)$ indicates whether the trajectory $\tau$ resulted in a correct final output. We adopt a trajectory-level success metric since it is non-trivial to get the success rate of each subquery, and the trajectory-level success rate provides a good estimate of a candidate action's effectiveness\footnote{We provide an empirical justification of the use of trajectory-level success rate as $PP$ in \Cref{sec:geb_validation}}.


\textbf{Collaboration Module Discovery.} In parallel to cost and performance estimations, successful trajectories are reflected upon by an LLM to identify recurring sequences of agent interactions that consistently contribute to successful task completion. These sequences are abstracted into reusable collaboration modules, which are added to the action space $\mathcal{A}'$ and refined through further rounds of self-play. Over successive iterations, this process yields a growing library of collaboration modules as well as reliable cost profiles for both modules and base agents.  

Through self-play reflection, \sys unifies cost and performance estimation, and collaboration module discovery in a single data-driven procedure, eliminating the need for manual engineering and enabling more effective budget-optimal planning.

\section{Experiments}
\label{sec:4}
To evaluate the effectiveness of \sys in multi-agent settings, we compare performance against baselines on agent benchmarks under strict budget constraints.

\subsection{Experiment Settings}

\stitle{Dataset.}
We evaluate \sys on three agent benchmarks: (1) GAIA \citep{mialon2023gaia}: a real-world QA benchmark that evaluates web browsing and general tool-use of language models; (2) BrowseComp-Plus \citep{chen2025browsecomp}: derived from BrowseComp \citep{wei2025browsecomp}, which measures agents' ability to browse the web using a fixed, curated corpus. (3) BigCodeBench \citep{zhuo2024bigcodebench}: a challenging coding benchmark consisting of function-level generation tasks with complex instructions. We use the Instruct subset.

\stitle{Agents and Collaboration Modules.} 
For each benchmark, we first instantiate a set of task-specialized agents. 
In GAIA, we utilize four agents, Search Agent, Browser Agent, Reasoning Agent, and Media Inspector Agent. In BrowseComp-Plus, we use three agents, Retriever Agent, Document Reader Agent, and Critic Agent. In BigCodeBench, we define two agents, Coding Agent and Verifier Agent.

To construct collaboration modules, we conduct five rounds of self-play reflection for each benchmark, with each producing one collaboration module. We utilize the first 30 questions in the benchmarks as a validation set to collect cost estimates and collaboration modules while evaluating the remaining questions. This process yields five modules for GAIA: \textit{interactive search and browse}, \textit{search then browse}, \textit{ensemble search}, \textit{two ensemble reasoning}, and \textit{three ensemble reasoning}, four modules for BrowseComp-Plus: \textit{interactive search}, \textit{ensemble interactive search}, \textit{interactive search then critic}, and \textit{ensemble interactive search then critic}, and three for BigCodeBench: \textit{code then verify}, \textit{ensemble code then verify}, and \textit{iterative code then verify}. Despite running five rounds of self-play reflection for BrowseComp-Plus and BigCodeBench, later rounds yield duplicates, resulting in fewer than five unique modules\footnote{The self-play reflection prompt and module descriptions are in \Cref{fig:selfplay_prompt} and \Cref{sec:agent_descriptions}}.

\stitle{Models.} 
We employ two model families: Claude \citep{anthropic_claude_overview} and Qwen3 \citep{qwen3}. Within the Claude family, we use \texttt{Claude-3.7-Sonnet} for the orchestrator, reasoning, critic, and media inspector agents, and \texttt{Claude-3.5-Haiku} for the rest. For the Qwen family, we use \texttt{Qwen3-32B} for all the agents.
For retrieving similar subqueries during performance prior calculation, we utilize \texttt{e5-large-v2}\footnote{https://huggingface.co/intfloat/e5-large-v2} embedding model to retrieve top-10 similar trajectories based on the cosine similarity.

\stitle{Budget Constraints.} 
For the unit of cost and budget, we compute the monetary cost of each action by multiplying its input and output token usage with the official token pricing from AWS Bedrock\footnote{The actual token price we used is in \Cref{sec:token_pricing}}. We evaluate performance under four budget settings for GAIA and BrowseComp-Plus and two for BigCodeBench. To determine suitable constraints for each benchmark and model, we first measure the trajectory costs from self-play reflection and take their average as the minimum budget. Larger budgets are then defined either by fixed increments or by exponential scaling. The resulting budget constraints are shown in \Cref{tab:main_perf}. 

\stitle{Metrics.}
We report results using Acc@$B$, which denotes the accuracy achieved under a budget $B$. This metric reflects the proportion of questions answered correctly while ensuring that the total token-based compute cost does not exceed $B$ for each query. Generations are stopped at $B$, after which the model directly generates an answer, following \cite{s1-2025}. Acc@$B$ enables fair comparison across different compute regimes and highlights how effectively each method converts its allotted budget into correct answers.

\stitle{Baselines.} 
We consider the following categories of baselines. (1) Fixed Agent Workflow: Optimized agent workflows that are fixed during the test time. We employ the most representative works, ADAS \citep{hu2024automated} and AFlow \citep{zhang2024aflow}. (2) ReAct with test-time scaling: We evaluate on the standard ReAct framework \citep{yao2023react} along with two well-established test-time scaling methods, Best-of-N \citep{brown2024large} and Iterative Verification \citep{self-refine2023}. (3) ReAct with Collaboration Modules: We incorporate collaboration modules into the ReAct loop and evaluate two settings: \texttt{Budget-Unaware}, where the orchestrator does not receive explicit budget information, and \texttt{Budget-Aware}, where the orchestrator is given explicit budget information in the prompt\footnote{We provide detailed implementation details in \Cref{sec:impl_details}}.

\begin{table*}[t]
\centering
\small
\setlength{\tabcolsep}{3pt}

\resizebox{0.95\linewidth}{!}{
\begin{tabular}{llcccccccccc}
\toprule
\multirow{2}{*}{\textbf{Claude Models}}&\multirow{2}{*}{Method}
& \multicolumn{4}{c}{GAIA}
& \multicolumn{4}{c}{BrowseComp-Plus}
& \multicolumn{2}{c}{BigCodeBench} \\
\cmidrule(lr){3-6} \cmidrule(lr){7-10} \cmidrule(lr){11-12}
&& Acc@0.2 & Acc@0.3 & Acc@0.4 & Acc@0.5
& Acc@0.2 & Acc@0.3 & Acc@0.4 & Acc@0.5
& Acc@0.025 & Acc@0.1 \\
\midrule
\multirow{2}{*}{\textit{Fixed Agent Workflow}}&ADAS
& 11.72 & 11.72 & 11.72 & 11.72
& 6.20 & 6.20 & 6.20 & 6.20
& 46.8 & 46.8 \\
&AFlow
& 12.96 & 12.96 & 12.96 & 12.96
& 5.42 & 5.42 & 5.42 & 5.42
& 46.1 & 46.1 \\
\midrule

\multirow{3}{*}{\makecell[l]{\textit{ReAct with}\\ \textit{Test-Time Scaling}}}&ReAct
& 35.80 & 35.80 & 35.80 & 35.80
& 24.33 & 24.66 & 24.66 & 24.66
& 45.8 & 45.8 \\
&\quad w/ BoN
& 35.80 & 35.80 & 37.03 & 37.65
& 24.33 & 24.80 & 24.80 & 24.80
& 46.1 & 46.2 \\
&\quad w/ IterVer
& 35.80 & 36.41 & 38.27 & 37.03
& 24.54 & 25.06 & 25.58 & 24.80
& 46.2 & 46.2 \\
\midrule
\textit{ReAct with }&Budget-Unaware
& 35.80 & 37.65 & 38.27 & 38.27
& 25.32 & 25.50 & 27.08 & 26.20
& 46.8 & 46.8 \\
\textit{Collaboration Modules}&Budget-Aware 
& 36.41 & 41.97 & 43.20 & 43.20
& 24.07 & 26.34 & 27.37 & 28.07
& 46.8 & 47.2 \\
\midrule
&\sys
& \textbf{38.89} & \textbf{44.44} & \textbf{46.91} & \textbf{48.15}
& \textbf{25.53} & \textbf{27.50} & \textbf{28.07} & \textbf{29.00}
& \textbf{47.0} & \textbf{48.2} \\
\bottomrule
\end{tabular}
}

\resizebox{0.95\linewidth}{!}{
\begin{tabular}{llcccccccccc}
\toprule
\multirow{2}{*}{\textbf{Qwen3-32B}}&\multirow{2}{*}{Method}
& \multicolumn{4}{c}{GAIA}
& \multicolumn{4}{c}{BrowseComp-Plus}
& \multicolumn{2}{c}{BigCodeBench} \\
\cmidrule(lr){3-6} \cmidrule(lr){7-10} \cmidrule(lr){11-12}
&& Acc@0.05 & Acc@0.1 & Acc@0.2 & Acc@0.3
& Acc@0.025 & Acc@0.05 & Acc@0.1 & Acc@0.2
& Acc@0.025 & Acc@0.1 \\
\midrule

\multirow{2}{*}{\textit{Fixed Agent Workflow}}&ADAS
& 4.93 & 4.93 & 4.93 & 4.93
& 2.84 & 2.84 & 2.84 & 2.84
& 46.0 & 46.0 \\
&AFlow
& 3.70 & 3.70 & 3.70 & 3.70
& 3.35 & 3.35 & 3.35 & 3.35
& 45.5 & 45.5 \\
\midrule
\multirow{3}{*}{\makecell[l]{\textit{ReAct with}\\ \textit{Test-Time Scaling}}}&ReAct
& 12.66 & 12.66 & 12.66 & 12.66
& 7.10 & 7.71 & 8.07 & 8.07
& 45.1 & 45.3 \\
&\quad w/ BoN
& 12.66 & 13.58 & 13.58 & 14.19
& 7.10 & 7.71 & 7.71 & 8.07
& 45.3 & 45.3 \\
&\quad w/ IterVer
& 12.66 & 14.19 & 16.00 & 16.67
& 8.07 & 14.72 & 15.24 & 16.14
& 45.3 & 45.3 \\
\midrule
\textit{ReAct with}&Budget-Unaware
& 15.33 & 16.00 & 16.00 & 16.00
& 10.36 & 16.14 & 17.46 & 17.46
& 45.5 & 45.5 \\
\textit{Collaboration Modules}&Budget-Aware 
& 11.33 & 13.33 & 14.00 & 14.00
& 16.98 & 16.74 & 16.74 & 16.74
& 45.5 & 45.8 \\
\midrule
&\sys
& \textbf{16.00} & \textbf{16.67} & \textbf{20.00} & \textbf{21.33}
& \textbf{18.07} & \textbf{18.92} & \textbf{19.42} & \textbf{20.72}
& \textbf{46.0} & \textbf{46.2} \\
\bottomrule
\end{tabular}
}
\caption{Evaluation results reported in Acc@$B$. Best-of-N and Iterative Verification are abbreviated as BoN and IterVer, respectively.}
\vspace{-2em}
\label{tab:main_perf}
\end{table*}

\subsection{Benchmark Results}
\label{sec:main_results}

\Cref{tab:main_perf} presents the experiment results of the baselines on the GAIA, BrowseComp-Plus, and BigCodeBench across different budget constraints. We highlight several key findings.

\textbf{(1) Fixed agent workflow baselines underperform on GAIA and BrowseComp-Plus, but outperform ReAct on BigCodeBench.} These benchmarks involve questions that require highly diverse workflows, causing optimized workflows in these methods to collapse into simple, generic pipelines. As a result, a single, static workflow struggles to adapt effectively across tasks. Moreover, because these approaches lack a mechanism for dynamically allocating compute, their performance remains largely unchanged as the available budget increases. However, on BigCodeBench, these baselines achieve competitive performance, and even surpass the ReAct baseline, suggesting that workflow optimization is particularly effective for coding tasks that can be solved in an end-to-end fashion. 

\textbf{(2) ReAct exhibits limited utilization of larger budgets.} Although adding Best-of-N or Iterative Verification partially mitigates this issue, these methods consume significantly more compute while yielding only minimal and inconsistent accuracy gains, which quickly plateau as the budget increases. This suggests that simply allocating more budget through standard test-time scaling does not guarantee improved performance.

\textbf{(3) Collaboration modules lead to clear and consistent improvements.} The Budget-Unaware variant already surpasses ReAct by a significant margin, confirming its effectiveness in using test-time compute. Moreover, while Budget-Aware baseline occasionally improves accuracy relative to non-budget-aware settings, its effect is inconsistent and in some cases plateaus at higher budgets. 

\textbf{(4) \sys achieves the strongest results across all budgets.} These gains highlight the value of dual-level planning, which enables the orchestrator to allocate test-time compute more effectively by balancing short-term decisions with long-horizon feasibility.

\subsection{Impact of Short-term and Long-term Planning}
To assess the contribution of each planning component in \sys, we evaluate them by isolating each component under \$0.2 and \$0.5 budgets using Claude models on GAIA benchmark. 
The results are shown in \Cref{tab:ablation_components}. Short-term planning alone provides a meaningful boost over vanilla ReAct, as it guides the orchestrator toward more informed 
\begin{wraptable}{r}{6.5cm}
\small
\centering


\setlength{\tabcolsep}{6pt}
\begin{tabular}{lcc}
\toprule
\textbf{Method} & \textbf{Acc@0.2} & \textbf{Acc@0.5} \\
\midrule
Only Short-term            & 36.41 & 44.44 \\
Only Long-term             & 37.65 & 46.91 \\
\textbf{Dual-level (Both)} & \textbf{38.89} & \textbf{48.15} \\
\bottomrule
\end{tabular}
\caption{Evaluation results of short-term and long-term planning on GAIA under \$0.2 and \$0.5 budget.}
\label{tab:ablation_components}

\end{wraptable}
local decisions and encourages the use of valuable collaboration modules. Long-term planning alone performs slightly better, particularly at higher budgets, since it enables the model to anticipate downstream computation and avoid prematurely committing to suboptimal trajectories. When combined, these complementary behaviors yield the strongest performance, underscoring the value of integrating both local and global planning signals for budget-optimal multi-agent coordination.

\begin{figure*}[t]
    \centering
    \begin{subfigure}[b]{0.45\textwidth}
        \centering
        \includegraphics[width=0.9\textwidth]{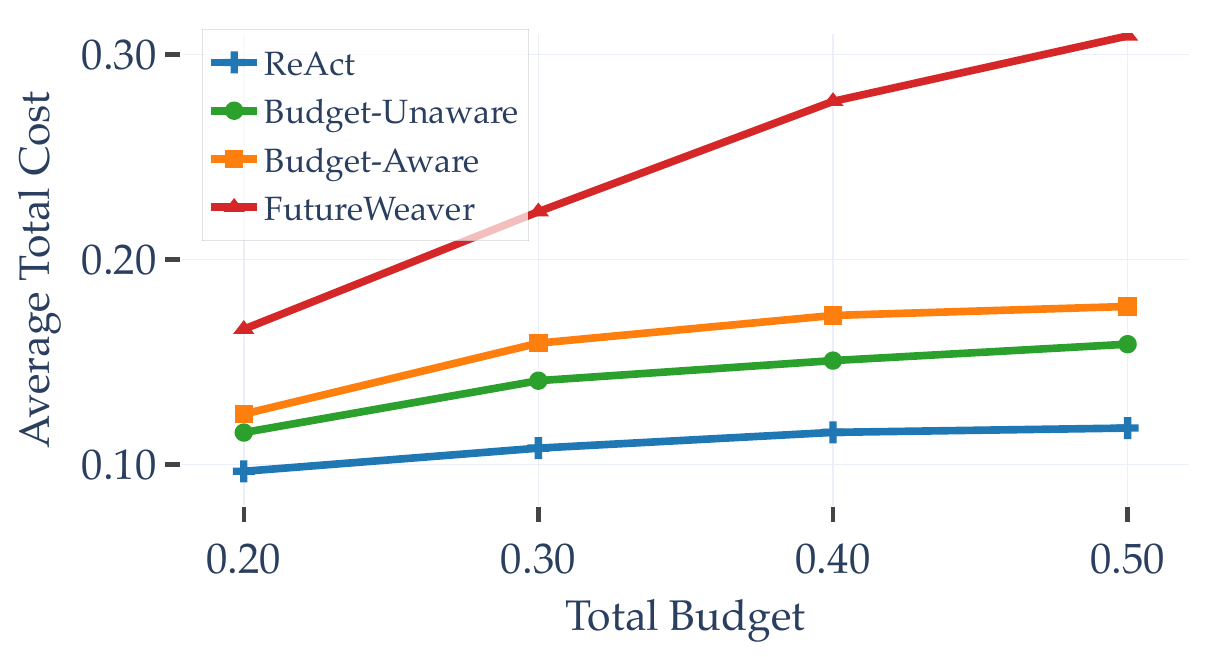}
        \label{fig:average_total_cost_claude}
    \end{subfigure}
    \begin{subfigure}[b]{0.45\textwidth}
        \centering
        \includegraphics[width=0.9\textwidth]{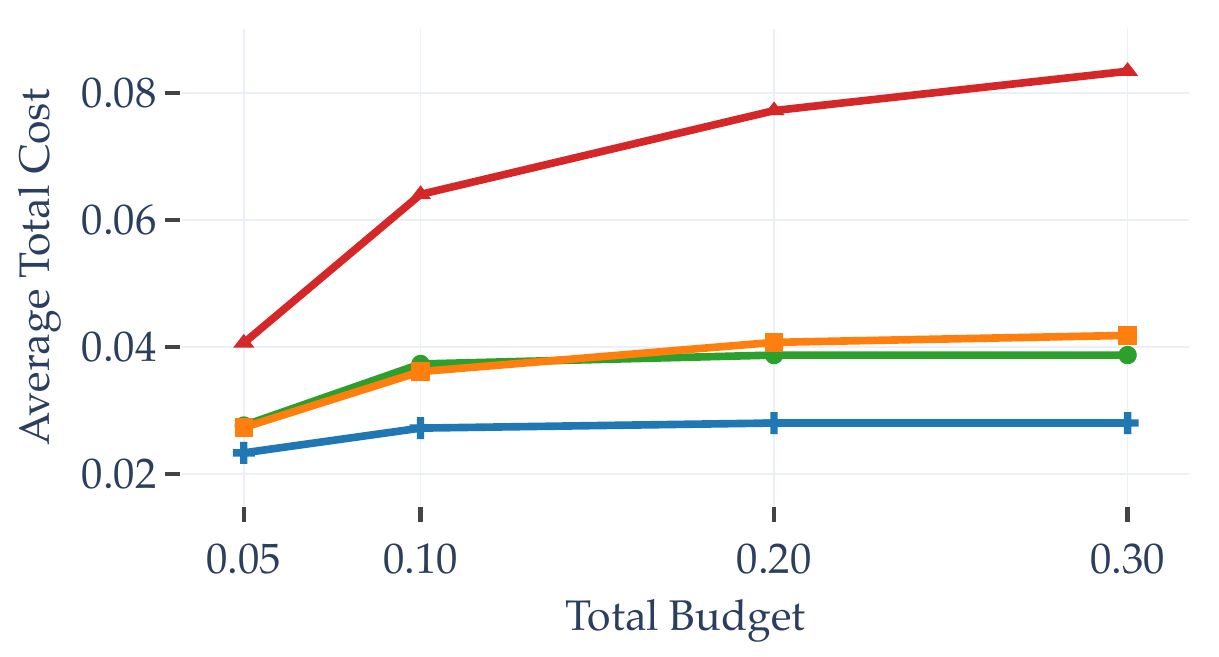}
        \label{fig:average_total_cost_qwen}
    \end{subfigure}
    
    \caption{Average total cost vs total budget on GAIA using Claude (left) and Qwen3 (right).}
    \label{fig:gaia_total_cost}
\vspace{-1em}
\end{figure*}

\subsection{Cost Utilization under Budget Constraints}

We further examine the average total cost incurred when running GAIA benchmark under different budget constraints, as reported in \Cref{fig:gaia_total_cost}. Across all three baselines, the available budget is consistently underutilized, with the total cost plateauing well below the specified constraint. In contrast, \sys achieves substantially higher cost utilization, which transfers into a performance enhancement. This finding highlights a fundamental challenge in multi-agent systems: even when sufficient budget is available, baseline strategies struggle to convert it into effective computation that drives performance gains. By leveraging dual-level planning, \sys allocates resources more aggressively when necessary, ensuring that budget headroom is effectively translated into higher performance.


\section{Conclusion}
We presented \sys, a general framework for budget-constrained optimization of test-time compute in multi-agent systems. \sys leverages collaboration modules as reusable abstractions of multi-agent coordination and employs a dual-level planning architecture that balances short-horizon execution with long-horizon speculation. Extensive experiments demonstrate that \sys consistently outperforms both standard baselines and those augmented with test-time scaling, achieving higher accuracy while utilizing resources more efficiently. These findings underscore the importance of structured collaboration and forward-looking planning for budget-constrained inference, and point toward a promising direction for building more adaptable and compute-efficient multi-agent systems.


\section*{Impact Statement}
This paper introduces \sys, a framework that advances machine learning by optimizing multi-agent test-time compute through automated collaboration discovery and dual-level planning. While this promotes resource efficiency, the increased autonomy in agent coordination requires developers and researchers to implement rigorous safety guardrails and bias monitoring to ensure these systems remain aligned with human values during autonomous deployment.

\bibliography{colm2026_conference}
\bibliographystyle{colm2026_conference}

\newpage

\appendix
\onecolumn

\section{Related Work}
\label{sec:related_work}
\stitle{Test-Time Scaling for LLMs}
Existing approaches to increasing test-time compute for LLMs can be broadly categorized into two paradigms, parallel scaling and sequential scaling. Parallel scaling, such as best-of-N sampling~\citep{brown2024large,snell2024scaling}, generates multiple candidate solutions in parallel and selects the best one using voting, confidence heuristics, or reward models~\citep{wang2022self,irvine2023rewarding}. In contrast, sequential scaling encourages iterative refinement via methods such as chain-of-thought prompting \citep{cot2022}, self-refinement \citep{self-refine2023,chen2023teaching,min2024imitate,lee2025evolving}, and verifier-guided revision \citep{gou2023critic,zhang-etal-2024-small}. 
While these methods offer promising directions for test-time scaling, they do not consider optimization under a strict compute budget. Recent works have explored budget-constrained settings \citep{s1-2025,han2025token,zheng2024budget,qiu2025co}, but they either focus on a single LLM setting or target system-level efficiency rather than scaling test-time computation to improve the performance. In contrast, our work focuses on budget-optimal compute allocation in a multi-agent setting, which is not addressed by these prior approaches.


\stitle{Designing and Optimizing Multi-Agent System}
Multi-agent systems (MAS) have recently gained traction as a way to structure complex tasks through the collaboration of specialized LLM-based agents. Early systems such as CAMEL~\citep{li2023camel}, AutoGen \citep{wu2024autogen}, and MetaGPT~\citep{hong2023metagpt} demonstrated the value of explicit role assignment and agent interaction, but relied heavily on manual configurations, including prompt engineering, agent profiling, and fixed communication protocols~\citep{qian2024scaling}. These limitations hinder their adaptability across domains and tasks.

In response, recent work has focused on automating various components of MAS design. Some methods treat agent functions as learnable policies \citep{zhang2024offline,zhang2024agent} or synthesize trajectories for offline agent optimization \citep{qiao-etal-2024-autoact}. Others expand the MAS search space to include prompts \citep{khattab2023dspy}, tools \citep{zhou2024symbolic}, workflows \citep{li2024autoflow}, and reasoning strategies \citep{shang2024agentsquare}. 
GPTSwarm~\citep{zhuge2024gptswarm} optimizes agent communication using policy gradients, and state-of-the-art systems like ADAS \citep{hu2024automated} and AFlow \citep{zhang2024aflow} perform full workflow optimization using search algorithms or LLM controllers. 
However, these systems aim to produce a single optimized configuration per task and do not support dynamic, inference-time planning under compute budgets, nor do they introduce reusable collaboration abstractions that enable adaptive, budget-efficient multi-agent coordination.

\section{Supplementary Details on Agents and Collaboration Modules}
\label{sec:agent_descriptions}
For the GAIA benchmark, we employ the following agents:
\begin{itemize}[leftmargin=*]
\item \textbf{Search Agent}: Given a search query, outputs the google search results
\item \textbf{Browser Agent}: Given one or more URLs, visits the pages and returns the content of webpages.
\item \textbf{Reasoning Agent}: Given a problem, performs multi-step reasoning and outputs a final solution.
\item \textbf{Media Inspector Agent}: Given an image or video link and a question, analyzes the media and answers the question.
\end{itemize}
In addition, we collect the following collaboration modules during self-play reflection:
\begin{itemize}[leftmargin=*]
\item \textbf{\texttt{interactive\_search\_and\_browse}}: A search agent and a browser agent share context and invoke each other interactively.
\item \textbf{\texttt{search\_then\_browse}}: First searches for URLs, then browses the retrieved pages for detailed content.
\item \textbf{\texttt{ensemble\_search}}: Generates three distinct queries, spawns three search agents in parallel, and aggregates their results.
\item \textbf{\texttt{two\_ensemble\_reasoning}}: Two reasoning agents independently produce reasoning paths, which are aggregated into a final answer.
\item \textbf{\texttt{three\_ensemble\_reasoning}}: Three reasoning agents independently produce reasoning paths, which are aggregated into a final answer.
\end{itemize}

For the BrowseComp-Plus benchmark, we employ the following agents:
\begin{itemize}[leftmargin=*]
\item \textbf{Retriever Agent}: Given a search query, retrieves the top-5 documents by semantic similarity and returns \texttt{(doc\_id, title, snippet)} for each.
\item \textbf{Document Reader Agent}: Given a \texttt{doc\_id}, fetches and returns the full document content.
\item \textbf{Critic Agent}: Given the current task state and available information, identifies missing information and recommends what to search for next.
\end{itemize}
We collect the following collaboration modules during self-play reflection:
\begin{itemize}[leftmargin=*]
\item \textbf{\texttt{interactive\_search}}: A search agent and a document reader agent share context and invoke each other interactively.
\item \textbf{\texttt{ensemble\_interactive\_search}}: Spawns three \texttt{interactive\_search} modules in parallel and aggregates their results.
\item \textbf{\texttt{interactive\_search\_then\_critic}}: First calls the \texttt{interactive\_search} module, then the critic agent evaluates the gathered information.
\item \textbf{\texttt{ensemble\_interactive\_search\_then\_critic}}: Spawns three \texttt{interactive\_search\_then\_critic} modules in parallel and aggregates their results.
\end{itemize}

For the BigCodeBench benchmark, we employ the following agents:
\begin{itemize}[leftmargin=*]
\item \textbf{Coding Agent}: Given a coding requirement, generates a solution code.
\item \textbf{Verifier Agent}: Given a coding requirement and a solution code, outputs the result of the verification. If the agent is provided with multiple solution codes, it verifies all the solution codes and outputs one solution code. 
\end{itemize}
We collect the following collaboration modules during self-play reflection:
\begin{itemize}[leftmargin=*]
\item \textbf{\texttt{code\_then\_verify}}: A coding agent generates the code solution and the downstream verifier agent outputs the result of the verification.
\item \textbf{\texttt{ensemble\_code\_then\_verify}}: Three coding agents generate codes in parallel, and the verifier agent verifies each code and selects the best code that will fulfill the requirements.
\item \textbf{\texttt{iterative\_code\_then\_verify}}: Initially run \texttt{code\_then\_verify}, then based on the verification, the code agent regenerates the code. We fix the iteration at 2. 
\end{itemize}

\section{Token Pricing for Cost Computation}
\label{sec:token_pricing}
We compute the monetary cost of each action by multiplying its input and output token usage with the official token pricing provided by AWS Bedrock\footnote{https://aws.amazon.com/bedrock/pricing/}. \Cref{tab:token_pricing} lists the exact token prices used for all models in our experiments. These values correspond to the pricing available at the time of experimentation.

Importantly, the Acc@B budget accounts for \emph{all} API calls made during inference, including those by the orchestrator for candidate generation, self-consistency computation, and long-horizon speculation. For Claude experiments, where \texttt{claude-3-7-sonnet-latest} serves as the orchestrator, this overhead accounts for roughly 15--20\% of total cost across all budget ranges. For Qwen experiments, the proportion is lower at roughly 7--15\% due to the lower token pricing. The dual-level planner is designed to be lightweight: short-term and long-term planning are generated jointly in a single LLM invocation per step, and self-consistency is computed over the sampled candidates without additional invocations.

\begin{table}[h!]
\centering
\begin{tabular}{lcc}
\toprule
\textbf{Model Name} & \textbf{Input Price} (\$/1K tok) & \textbf{Output Price} (\$/1K tok) \\
\midrule
\texttt{claude-3-5-haiku-latest}   & \$0.0008 & \$0.004 \\
\texttt{claude-3-7-sonnet-latest}  & \$0.003 & \$0.015 \\
\texttt{qwen3-32b}                            & \$0.0007 & \$0.0028 \\
\bottomrule
\end{tabular}
\caption{Token prices used for calculating costs and budgets.}
\label{tab:token_pricing}
\end{table}

\section{Reliability of Mean-Based Cost Estimation}
\label{sec:reliabilty_mean_cost}
\begin{table}[h]
\centering
\small
\begin{tabular}{lccc}
\toprule
\textbf{Module / Agent} & \textbf{Self-Play Est.\ (\$)} & \textbf{Deploy Mean $\pm$ Std (\$)} & \textbf{Gap (\%)} \\
\midrule
search\_agent                  & 0.026 & 0.029 $\pm$ 0.007 & +8.8  \\
browser\_agent                 & 0.045 & 0.053 $\pm$ 0.035 & +17.8 \\
reasoning\_agent               & 0.026 & 0.031 $\pm$ 0.009 & +19.5 \\
media\_inspector\_agent        & 0.022 & 0.024 $\pm$ 0.008 & +7.3  \\
ensemble\_search               & 0.034 & 0.038 $\pm$ 0.013 & +11.8 \\
interactive\_search\_and\_browse & 0.060 & 0.052 $\pm$ 0.037 & $-$14.2 \\
search\_then\_browse            & 0.068 & 0.061 $\pm$ 0.036 & $-$11.1 \\
two\_ensemble\_reasoning       & 0.085 & 0.105 $\pm$ 0.029 & +23.5 \\
three\_ensemble\_reasoning     & 0.121 & 0.142 $\pm$ 0.039 & +17.4 \\
\bottomrule
\end{tabular}
\caption{Self-play cost estimates vs.\ actual deployment costs on GAIA (Claude models).}
\label{tab:cost-gap}
\end{table}
 
Since \sys approximates module costs by averaging over self-play trajectories, a natural concern is whether these estimates remain accurate at deployment time. \Cref{tab:cost-gap} reports the self-play estimates alongside the actual deployment mean and standard deviation for each agent and collaboration module on GAIA with Claude models.
 
Estimation gaps range from 7\% to 24\%, confirming that mean-based estimation is reliable in practice. Module costs are dominated by the structurally fixed number of LLM invocations, which limits variance. Browsing-related and reasoning-heavy modules exhibit higher standard deviations due to variable content lengths, but the gaps remain manageable. Furthermore, the dual-level planner provides a natural correction mechanism: it re-evaluates budget feasibility at every step using the \emph{actual} remaining budget, so estimation errors are corrected dynamically rather than compounding over the trajectory.

\section{Implementation Details}
\label{sec:impl_details}
\subsection{Baselines}
Here, we provide implementation details for the baselines. For the best-of-N baseline, we set $N=3$, but if the budget is exhausted before completing all three attempts, we use the available attempts and apply self-consistency on the answers from the attempts to produce the final answer. For the iterative verification baseline, once an initial answer is generated, any remaining budget is used to prompt the orchestrator to re-examine the trajectory and refine the solution until the budget is fully consumed.

\subsection{Dual-level Planning Architecture}
Short-term and long-term planning are not computed in separate LLM calls. Instead, they are generated jointly in a single invocation to reduce overhead. Given the current task state, execution history, and the previously successful long-term plan from the prior round, the orchestrator produces multiple candidate next actions (short-term planning), each accompanied by a plausible continuation sequence of future modules (long-term planning). For example, a candidate might propose ``interactive browse'' as the next action, with a continuation of ``$\rightarrow$ ensemble reasoning $\rightarrow$ critic'' as the speculative long-term plan.
 
Each candidate's continuation sequence is evaluated against the remaining budget using the cost estimates from self-play reflection. Sequences whose estimated cumulative cost exceeds the remaining budget are filtered out, and the proportion of budget-feasible continuations determines the feasibility score $h(\cdot)$. The feasible long-term plans are then carried forward to the next round, where they condition both the next step's candidate generation and its speculative continuations. This creates a feedback loop in which successful long-term plans progressively guide future decisions.
 
Crucially, accuracy of intermediate state prediction is not required. The speculative trajectories serve as a budget feasibility check: they estimate whether a given next action leaves enough budget for a plausible completion, not what the completion will produce. This is why module-level granularity suffices and the entire procedure remains lightweight.


\section{Validation of Trajectory-Level Success as a Performance Prior}
\label{sec:geb_validation}

In this section, we provide an empirical justification of using trajectory-level success rate as the Performance Prior ($PP$) in short-term planning. We conducted an ablation study comparing our trajectory-level estimate against a fine-grained, subquery-level success metric.

\paragraph{Experimental Setup} 
In this setup, we employ \texttt{gpt-o3} \citep{openai_o3_o4mini_2025} as an \textit{LLM-as-a-judge} to evaluate the success of individual subqueries within the collected trajectories during self-play reflection ($\mathcal{D}_{ref}$). For each invocation $(a, v)$ in a trajectory, the judge was provided with the subtask $v$, the agent's output $o$, and the ground-truth final answer to determine if the specific action correctly contributed to the task progress. This fine-grained $PP$ was then integrated into the short-term planning phase. We tested on GAIA and BrowseComp-Plus using Claude models and used Acc@0.5 as the performance metrics.

\paragraph{Results and Discussion}
As shown in Table~\ref{tab:finegrained_comparison}, utilizing a subquery-level reward signal does not necessarily lead to a better performance in GAIA and BrowseComp-Plus. We attribute this to the high degree of coupling in multi-agent workflows; a correct subquery is only valuable if the subsequent agents are capable of utilizing its output. These results suggest that trajectory-level success provides a sufficiently robust and computationally efficient estimate for performance estimation without the need for more fine-grained success metrics.

\begin{table}[h]
\centering
\small
\begin{tabular}{lcc}
\toprule
\textbf{$PP$ Formulation} & \textbf{GAIA (Acc@0.5)} & \textbf{BrowseComp-Plus (Acc@0.5)} \\
\midrule
Trajectory-level (Ours) & 48.15 & 29.00 \\
Subquery-level (GPT-o3 Judge) & 48.32 & 28.87 \\
\bottomrule
\end{tabular}
\vspace{0.6em}
\caption{Comparison of trajectory-level success vs. fine-grained subquery-level judging for $PP$ calculation. The fine-grained approach provides only marginal gains despite significantly higher computational overhead.}
\label{tab:finegrained_comparison}
\end{table}

\section{Dual-Level Planning Algorithm}
We provide the algorithm for dual-level planning in \Cref{alg:dual_planning}.

\newpage

\begin{tcblisting}{%
  enhanced,
  breakable,
  colback=lightgray,
  colframe=black!30,
  boxrule=0.3pt,
  arc=2pt,
  auto outer arc,
  width=\textwidth,
  listing only,
  listing engine=listings,
  left=2mm,right=2mm,top=1mm,bottom=1mm,
  fontupper=\small\ttfamily
}
You are given a set of successful execution trajectories produced by a multi-agent system.
Each trajectory contains the sequence of agents invoked with a subtask, and their intermediate outputs

Your goal is to analyze these trajectories to identify recurring agent interactions that frequently appear in successful trajectories.
Such patterns may include:
- Sequential workflows 
- Parallel workflows 
- Verification or Refinement

For each recurring pattern you identify:
1. Describe the workflow in plain language.
2. Specify the agents involved and their roles.
3. Implement a class that executes this workflow using these agents 
4. Explain why this pattern is effective, referring to the trajectory evidence.

Here are some examples of the extracted pattern:
{few_shot_demonstrations}

Here are the patterns that are already observed. Avoid outputting duplicated patterns. 
{collected_collaboration_modules}
\end{tcblisting}
\captionof{figure}{Prompt used for self-play reflection to induce collaboration modules from successful trajectories.}
\label{fig:selfplay_prompt}

\begin{algorithm}[t]
\caption{Dual-Level Planning}
\label{alg:dual_planning}
\KwIn{Policy $\pi$, set of agents/modules $\mathcal{A}'$, exploration buffer $\mathcal{D}_{\text{exp}}$, initial history $\mathcal{H}_0$, total budget $B$, max steps $T_{\max}$}
\KwOut{Final output and execution trace}

$t \gets 0$\; $\mathcal{H}_t \gets \mathcal{H}_0$\; $B_t \gets B$\; $\mathcal{T}_{\text{feasible},0} \gets \emptyset$\;

\While{$t < T_{\max}$ \textbf{and} $B_t > 0$ \textbf{and} \textbf{not} \textsc{Solved}$(\mathcal{H}_t)$}{
  Sample candidate actions:
  \[
  \mathcal{C}_t = \{\alpha_t^{(1)}, \dots, \alpha_t^{(K)}\} \sim \pi(\cdot \mid \mathcal{H}_t, \mathcal{T}_{\text{feasible},t-1})
  \]
  \tcp{Let $\phi(\alpha)=a$ and $\psi(\alpha)=v$ extract components}

  \For{$i=1$ \KwTo $K$}{
    Compute Self-Consistency:
    \[
    SC(\alpha_t^{(i)}) = \frac{1}{K}\sum_{k=1}^{K}\mathbf{1}\!\left[\phi(\alpha_t^{(k)})=\phi(\alpha_t^{(i)})\right]
    \]
    Retrieve similar subqueries $S_{top10}(\psi(\alpha_t^{(i)}))$ from $\mathcal{D}_{\text{exp}}$ \;
    Compute $PP$ from trajectories $\mathcal{T}(S_{top10}(\psi(\alpha_t^{(i)}))$:
    \begin{equation}
PP(\alpha_t^{(i)}) = \frac{\sum_{\tau \in \mathcal{T}(S_{top10}(\psi(\alpha_t^{(i)}))} \mathbf{1}[Success(\tau)]}{|\mathcal{T}(S_{top10}(\psi(\alpha_t^{(i)}))|}
\end{equation}
    Compute short-term gain:
    \[
    g(\alpha_t^{(i)}) = \sqrt{SC(\alpha_t^{(i)}) \cdot PP(\alpha_t^{(i)})}
    \]
  }
  
  \For{$i=1$ \KwTo $K$}{
    Generate speculative trajectories $\mathcal{T}_t(\alpha_t^{(i)})$ beginning with $\alpha_t^{(i)}$\;
    Define feasible set:
    \[
    \mathcal{T}_{\text{feasible},t}(\alpha_t^{(i)}) = \big\{\tau \in \mathcal{T}_t(\alpha_t^{(i)}) \;\big|\; \text{cost}(\tau) \le B_t \big\}
    \]
    Compute long-term feasibility score:
    \[
    h(\alpha_t^{(i)}) = \frac{|\mathcal{T}_{\text{feasible},t}(\alpha_t^{(i)})|}{\sum_{r=1}^{K}|\mathcal{T}_{\text{feasible},t}(\alpha_t^{(r)})|}
    \]
  }
  
  Compute total utility $f(\alpha_t^{(i)}) = g(\alpha_t^{(i)}) + h(\alpha_t^{(i)})$ for $i=1,\dots,K$\;
  \[
  \alpha_t^\star = \arg\max_{\alpha \in \mathcal{C}_t} f(\alpha)
  \]
  
  Execute $\alpha_t^\star=(a_t^\star,v_t^\star)$ to obtain output $o_t$\;
  Update state: $\mathcal{H}_{t+1} \gets \mathcal{H}_t \cup \{(v_t^\star, o_t)\}$\;
  Update budget: $B_{t+1} \gets B_t - \text{cost}(\alpha_t^\star)$\;
  $t \gets t+1$\;
}
\Return{Final output derived from $\mathcal{H}_t$}
\end{algorithm}

\end{document}